\documentclass[preprint]{article} 
\usepackage{aaai2027}  
\usepackage[hyphens]{url}  
\usepackage{graphicx} 
\usepackage{natbib}  
\usepackage{caption} 
\usepackage{algorithm}
\usepackage{algorithmic}

\usepackage{newfloat}
\usepackage{listings}
\DeclareCaptionStyle{ruled}{labelfont=normalfont,labelsep=colon,strut=off} 
\floatstyle{ruled}
\newfloat{listing}{tb}{lst}{}
\floatname{listing}{Listing}

\usepackage{booktabs}

\newcommand{\ours}{SLED} 

\title{SLED: Scalable Location Encoding via Distillation}
\author {
    Kevin Lane\textsuperscript{\rm 1},
    Zhongying Wang\textsuperscript{\rm 1},
    Esther Rolf\textsuperscript{\rm 1},
    Morteza Karimzadeh\textsuperscript{\rm 1}
}
\affiliations {
    \textsuperscript{\rm 1}University of Colorado Boulder\\
    Kevin.Lane@colorado.edu, Zhongying.Wang@colorado.edu, Esther.Rolf@colorado.edu, Karimzadeh@colorado.edu
}

\begin{document}

\maketitle

\begin{abstract}
The plethora of readily available geospatial data offers exciting opportunities to learn high quality representations of the planet, but the sheer size of the Earth Observations (EO), differing modalities, and different sensor types pose significant challenges in doing so. Location encoders have emerged as an efficient way of compressing EOs into location-specific embeddings.  However, current state-of-the-art location encoders rely on computationally expensive CLIP-style frameworks that require large batch sizes in the 16K--32K range, suffer from false negative samples, and scale poorly with additional modalities. We introduce the Scalable Location Encoder via Distillation (\ours), a distillation-based location encoder that uses geospatial location as a binding modality to pretrain location encoders with any modality of geospatial data. The resulting location encoder framework is lightweight, modular, and can flexibly incorporate multiple modes, while eliminating the need for spatiotemporal coregistration of samples. \ours{} is performant with batch sizes as small as 128, enabling pretraining at a fraction of the runtime and compute costs of current state-of-the-art models.  We demonstrate our approach by pretraining unimodal and multimodal \ours{} models on Sentinel-1, Sentinel-2, and Landsat imagery.   We show that both unimodal and multimodal \ours{} models keep pace with or outperform existing approaches on a diverse set of 19 human-centric benchmark tasks and explore the benefits of using additional modes in pretraining. 
\end{abstract}

\begin{links}
  \link{Code \& Pip Install}{https://github.com/geohai/sled}
  \link{Datasets \& Weights}{https://huggingface.co/geohai}
\end{links}

\section{Introduction}

Models capable of learning general purpose representations of Earth Observations (EO) have become increasingly popular \cite{maiNextGenerationGeospatial2025}.  Embeddings generated from one such geospatial foundation model (GeoFM) can be leveraged on a variety of downstream geospatial tasks \cite{rolfGeneralizableAccessibleApproach2021}, including flood plain mapping \cite{bonafiliaSen1Floods11GeoreferencedDataset2020}, air pollution monitoring \cite{karimzadeh2025performance}, land cover classification \cite{christie2018functional}, and many others \cite{lacoste2023geo}. 

Despite the efficacy of GeoFM embeddings, using them can be challenging.  In order to generate embeddings, users have to first acquire and process data of a particular modality--or modalities, in the case of multimodal GeoFMs such as \citet{astrucOmniSatSelfsupervisedModality2025, danish2025terrafm, fuller2023croma, jakubik2025terramind, xiong2024neural}--and run inference on them. For perspective, the popular Sentinel-2 satellite constellation produces 1.7 TB of data per day \citep{Sudmanns02072020} and has been in orbit for over a decade.  The sheer size means that downloading data and passing samples through pretrained GeoFM models often requires significant time, storage, and computational resources. 

\begin{figure*}[t!]
\centering
\begin{minipage}{0.45\textwidth}
  \centering
  \includegraphics[width=0.9\linewidth]{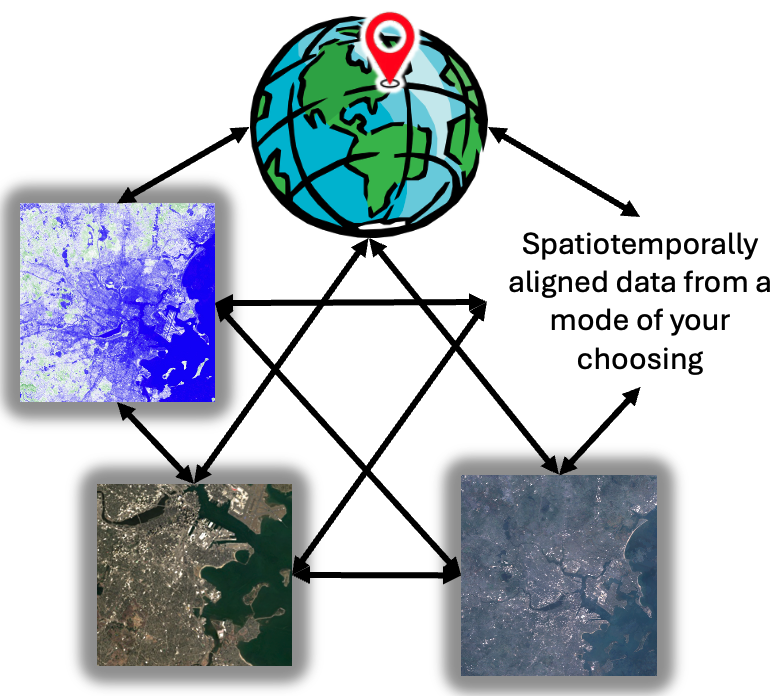}
  \caption{Multimodal location encoding without location as a binding modality.  All modes must be spatiotemporally coregistered per sample.  The sample above is of Boston, MA, USA, with samples from Sentinel-1, Sentinel-2, and Landsat 9 in Fall, 2024.}
  \label{fig:unigeoclip_setup}
\end{minipage}%
\hfill
\begin{minipage}{0.45\textwidth}
  \centering
  \includegraphics[width=0.9\linewidth]{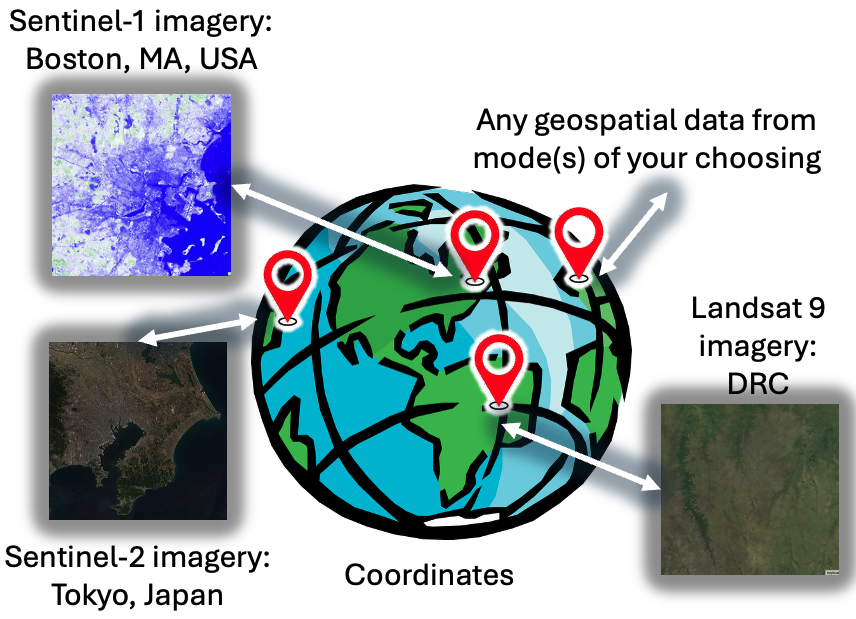}
  \caption{Multimodal location encoding with location as a binding modality.  We leverage the same satellite sensors present in Fig \ref{fig:unigeoclip_setup}, but can do so via independent samples.  Note the lack of requirement for spatiotemporal coregistration between modes.}
  \label{fig:ours_setup}
\end{minipage}
\end{figure*}

Given these limitations, a type of GeoFM called a location encoder has emerged \cite{maiCSPSelfSupervisedContrastive2023, klemmerSatCLIPGlobalGeneralPurpose2025, vivancocepedaGeoCLIPClipInspiredAlignment2023, dollinger2025climplicit}, which treats geospatial position, in the form of latitude and longitude, as its own modality.  These models train an encoder to learn representations of geospatial position by feeding the location encoder latitude and longitude, then contrasting the resulting location embeddings with EO embeddings associated with that location extracted from pretrained encoders.  Thus, location encoders learn how to parametrically generate representations of \textit{any} geospatial location, and are capable of interpolating to spatial gaps within their training samples \cite{klemmerSatCLIPGlobalGeneralPurpose2025}.  Location encoders do not require (but optionally allow the appendage of) an input image (or other modes) at the time of inference \cite{klemmerSatCLIPGlobalGeneralPurpose2025, vivancocepedaGeoCLIPClipInspiredAlignment2023, maiCSPSelfSupervisedContrastive2023, astrucUNIGEOCLIPUnifiedGeospatial2026, liuGAIRImprovingMultimodal2025}.  Given the inherent geospatial nature of many tasks, location embeddings can be easily leveraged in existing workflows and can dramatically improve performance on classification and regression tasks alike \cite{wu2024torchspatial, karimzadeh2025performance}.

While existing location encoders represent a significant step forward in accessible generation of embeddings, they generally remain computationally expensive to train.  Almost all existing location encoders \cite{klemmerSatCLIPGlobalGeneralPurpose2025, maiCSPSelfSupervisedContrastive2023, vivancocepedaGeoCLIPClipInspiredAlignment2023, astrucUNIGEOCLIPUnifiedGeospatial2026, liuGAIRImprovingMultimodal2025} rely on the InfoNCE loss \cite{radford2021learning} between location embeddings and modality-specific embeddings.  This requires large batch sizes \cite{klemmerSatCLIPGlobalGeneralPurpose2025}, often in the range of 16k-32K, which can create compute bottlenecks.  InfoNCE loss also suffers from false negatives, where similar samples within a batch are treated the same as disparate samples. Given the inherent similarity of spatially-nearby locations \cite{waters2017tobler}, this is a particularly undesirable property for location encoders. Addressing false negatives remains an active area of research for CLIP-style models \cite{auh2026mitigating, huynh2022boosting}.

Despite their promising qualities, existing location encoders either do not leverage multimodal geospatial data or do so in a way that constrains their training dataset.  Many current location encoders are trained with a single modality of geospatial data,  perhaps due to the computational cost of adding additional modalities and the required batch size for contrastive training. Those that train against more than one mode do so via contrasting samples across all data, as seen in Figure \ref{fig:unigeoclip_setup}.  This requires costly cross-modal coregistration for every sample and excludes any training samples that do not have all modes of information available at a given location.  It also introduces subjectivity into the training process, as samples are rarely perfectly spatiotemporally aligned across modes and thus require some spatiotemporal buffering \cite{stewart2023ssl4eo}, but the degree to which that is done varies from dataset to dataset. 

We propose a novel distillation-style framework, the Scalable Location Encoder via Distillation (\ours), for distilling location embeddings from (vision) GeoFMs.  Our framework's computational complexity scales linearly with batch size, avoids false negative issues present in existing models, and exhibits strong performance with batch sizes as small as 128.  We also take inspiration from ImageBind \cite{girdhar2023imagebind} and \cite{klemmer2025earth} to treat location as a binding modality when training multimodal versions of \ours.  This allows us to leverage new modalities while avoiding the subjective and costly process of spatiotemporally coregistering samples from different modalities, as demonstrated in both Figure \ref{fig:unigeoclip_setup} and Figure \ref{fig:ours_setup}.

Our core contributions are as follows:
\begin{enumerate}
    \item We demonstrate that it is possible to distill unimodal and multimodal EOs into performant location embeddings via Mean Squared Error (MSE) Loss with batch sizes as small as 128 samples.
    \item We show that \ours's distillation framework speeds up pretraining by up to 47x when compared to pretraining times for existing state-of-the-art (SOTA) location encoders.
    \item We make pretrained versions of \ours{} and our pretraining framework available via a pip module \texttt{sled-geo}, which is designed to work with \texttt{pytorch-lightning}.
    \item We release a location encoder benchmark dataset of climate-related variables and a separate uniform-at-random Sentinel-1 location encoder pretraining dataset. 
\end{enumerate}

\section{Related Work}
\begin{figure*}[t!]
  \centering
  \includegraphics[width=0.9\linewidth]{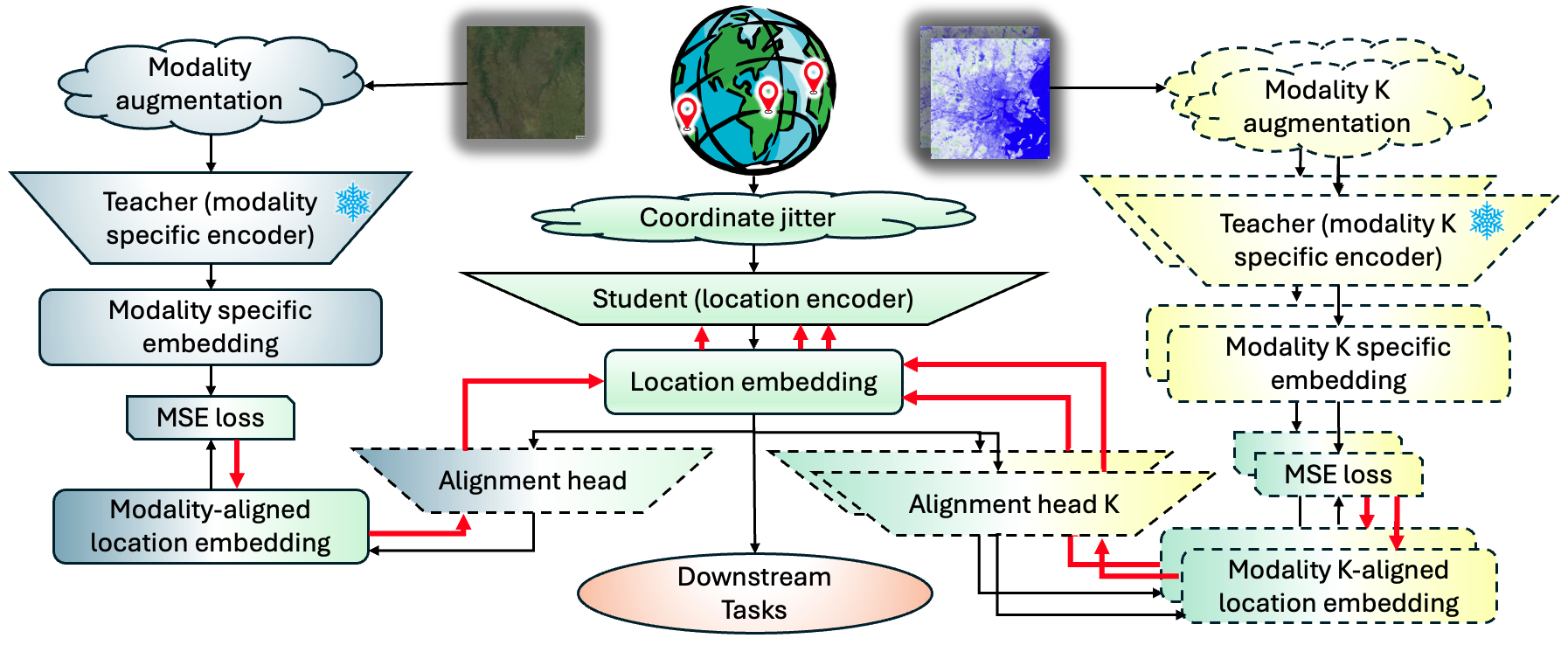}
  \caption{Diagram of \ours's training framework with $1$ to $K$ modes.  The current model outlines when $K=3$.  Components outlined with a dashed line are only present when training with $2$ or more modes.  For unimodal training, no alignment head is present.  The red arrow in bold denotes back-propagation.}
  \label{fig:framework_diagram}
\end{figure*}
Among the various GeoFMs developed, two broad categories have emerged:
\begin{enumerate}
    \item Foundation models trained to generate embeddings from one or more of their geospatial training modalities, such as Sentinel-2. Examples include TerraFM \cite{danish2025terrafm}, FG-MAE \cite{wang2024feature}, OmniSat \cite{astrucOmniSatSelfsupervisedModality2025}, DOFA \cite{xiong2024neural}, DeCUR \cite{wang2023decur}, CROMA \cite{fuller2023croma}, and GFM \cite{mendieta2023towards}. These models require access to the input modality (or modalities) at downstream inference, follow a myriad of different designs, and have been surveyed in \cite{lane2025genealogy} and \cite{lu2025vision_fm}.
    
    \item Location encoder models trained to generate embeddings directly from latitude and longitude, without requiring access to the original input modality. Prior work typically use contrastive learning against geospatial representations derived from data at those locations. Examples include CSP \cite{maiCSPSelfSupervisedContrastive2023}, SatCLIP \cite{klemmerSatCLIPGlobalGeneralPurpose2025}, GeoCLIP \cite{vivancocepedaGeoCLIPClipInspiredAlignment2023}, GAIR \cite{liuGAIRImprovingMultimodal2025}, CLIMPLICIT \cite{dollinger2025climplicit}, and UniGeoCLIP \cite{astrucUNIGEOCLIPUnifiedGeospatial2026}. Of these, only GAIR \cite{liuGAIRImprovingMultimodal2025} and UniGeoCLIP \cite{astrucUNIGEOCLIPUnifiedGeospatial2026} are pretrained with multiple data modalities.  Both rely on InfoNCE loss and require spatiotemporal alignment of their training modalities.  CLIMPLICIT \cite{dollinger2025climplicit} is the only model present that does not rely on InfoNCE loss, rather using MSE loss against a vector of climatic variables.
\end{enumerate}

The first type of model requires relatively complete spatial coverage of input modalities to generate embeddings indexed to latitude and longitude, as in MOSAIKS \cite{rolfGeneralizableAccessibleApproach2021}, AlphaEarth \cite{brown2025alphaearth}, and OlmoEarth \cite{herzog2025olmoearth}. In contrast, location encoders learn spatially continuous representations for \textit{any} location without requiring a database of indexed embeddings wherever input modalities were available. This is partially enabled by the positional encoding component of the location encoder.

Positional encodings map geographic coordinates to (learnable) high-dimensional representations \cite{mai2022review}.  Recent work builds on concepts of periodic activation functions \cite{sitzmann2020implicit} to enable learning high-frequency spatial signals, by incorporating Earth-aware basis functions, such as spherical harmonics, in conjunction with sinusoidal neural networks to produce continuous, multi-scale embeddings of geographic coordinates \cite{russwurmgeographic}. While they use similar loss functions, existing work on pretrained location encoders vary in  their position encoding strategy.  SatCLIP \cite{klemmerSatCLIPGlobalGeneralPurpose2025} utilizes spherical harmonics \cite{russwurmgeographic}, while GeoCLIP \cite{vivancocepedaGeoCLIPClipInspiredAlignment2023}, GAIR \cite{liuGAIRImprovingMultimodal2025}, and UniGeoCLIP \cite{astrucUNIGEOCLIPUnifiedGeospatial2026} use some form of Random Fourier Features (RFF) \cite{tancik2020fourier}.  

Location encoders differ in training modalities as well.  SatCLIP was trained on spatially uniform distributed Sentinel-2 imagery, while GeoCLIP was trained on ground-level Flickr imagery that is primarily available in North America and Europe.  GAIR \cite{liuGAIRImprovingMultimodal2025} incorporated Local Implicit Image Function \cite{chen2021learning} to fuse embeddings from point-based ground level imagery \cite{houGlobalStreetscapesComprehensive2024} into large footprints of Sentinel-2 imagery. UniGeoCLIP \cite{astrucUNIGEOCLIPUnifiedGeospatial2026} performs multimodal training between location and 4 other modes, but does so on proprietary data.  The multimodal contrastive methodology for GAIR and UniGeoCLIP can broadly be described by Figure \ref{fig:unigeoclip_setup}.  

There is also evidence to suggest that concatenation of representations from complementary models, such as SatCLIP and AlphaEarth, improves generalizable downstream performance when compared to using one model's representations \cite{van2026better}.  However, the optimal manner of fusing representations together and choosing complementary models is not yet well understood.

\ours{} represents the first location encoder to learn location embeddings via distillation and the first to do so while using location as a binding modality, which enables a new way to fuse complementary EO embeddings.

\section{Methodology}
\begin{figure*}[t!]
\centering
  \includegraphics[width=0.9\linewidth]{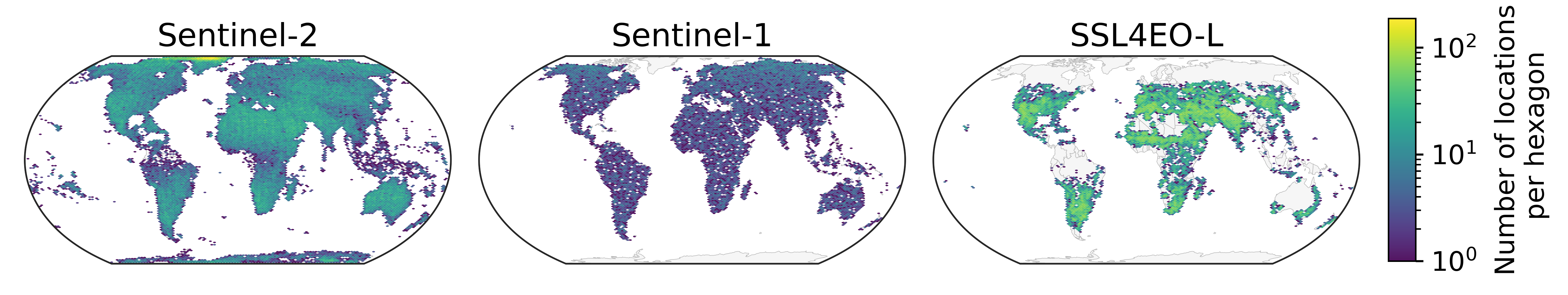}
  \caption{Spatial density of the pretraining datasets used in this work.}
  \label{fig:dataset_spatial_distribution}
\end{figure*}

\subsection{Framework}
A schematic overview of our framework is given in Figure \ref{fig:framework_diagram}.
When training with a single modality, we simply compute MSE loss between our location embeddings and modality-specific embeddings from a frozen teacher.  When training with more than one modality, we take inspiration from ImageBind \cite{girdhar2023imagebind} and treat position as the single \textit{binding modality}.  We introduce $K$ parallel linear layers on top of the position encoder to serve as alignment heads, where K is the number of modes.  These alignment heads serve two main purposes: (1) They decouple the size of the position encoder's latent space from the sizes of the teacher encoders' latent spaces, improving model flexibility, (2) They improve model convergence while avoiding embedding collapse.

With each additional data modality, an additional frozen modality-specific teacher and trainable alignment head are added to the model.  Each sample has modality-specific data, latitude, and longitude.  When training, all samples are fed through the online student location encoder and their corresponding modality-specific teacher encoder.  In this manner, every sample fires the student encoder and its respective modality-specific teacher encoder.  

For the teacher encoders, we use the ViT-L version of TerraFM \cite{danish2025terrafm} for Sentinel-2, the ViT-L version of FG-MAE \cite{wang2024feature} for Sentinel-1, and the ViT-B version of MoCo \cite{he2020momentum} pretrained on Landsat imagery for Landsat 8/9.  FG-MAE and MoCo are obtained through the TorchGeo python package \cite{stewart2024torchgeo}.  Additional details may be found in the supplemental materials.

\subsection{Data}
Figure \ref{fig:dataset_spatial_distribution} shows the spatial distribution of our training datasets. In each modality, a sample consists of an image and a corresponding latitude and longitude. \ours\ readily allows for incorporating multiple modalities. When training \ours\ for our experiments, we distill information from up to three distinct modalities of satellite imagery:
\begin{itemize}
    \item Sentinel-2 level 2 A: 100K samples of 12-band, $256 \times 256$ resolution multi-spectral imagery from the S2-100K dataset originally generated by \cite{klemmerSatCLIPGlobalGeneralPurpose2025}.
    \item Landsat 8/9 OLI SR: 250K samples of 7-band, $224 \times 224$ multi-spectral imagery from the SSL4EO-L dataset \cite{stewart2023ssl4eo}. The SSL4EO-L dataset contains 250K locations, each with 4 images corresponding to the seasons.  In order to avoid temporal overlap, we randomly choose one image per location to use in our  dataset. 
    \item Sentinel-1: 20K samples of $224 \times 224$ resolution C-band Synthetic Aperture Radar (SAR) imagery with VH/VV polarization sampled in uniform at random (UAR) spatial distribution on landmasses.   We created this dataset via the Microsoft Planetary STAC API and have made it available for use.
\end{itemize}

We selected these three freely-available satellite data modalities as widely used representative EO that capture complementary information across spectral, spatial, and sensing modalities. Optical multi-spectral Sentinel-2 and Landsat satellites provide globally consistent observations, and Sentinel-1 provides all-weather microwave measurements, allowing for experimentation for various tasks. It is worth noting, however, that available benchmarks, including those that we use in this paper, are primarily based on tasks where optical imagery excels. 

\subsection{Pretraining Specifications}
To isolate the impact of multimodal training, all models are pretrained on 100K samples in total, meaning in multimodal training, only subsets of each satellite dataset are used to keep the total training samples constant. We train with early stopping, randomly selecting 10\% of our training dataset to use as a validation set, with a minimum validation loss improvement of 0.001 and a patience of 10 epochs.   We limit pretraining to 150 epochs.  When pretraining with multiple modes, the distribution of our validation set is proportional to the overall distribution of our training dataset (i.e., if we pretrain with 50k Sentinel-2 images and 50k Landsat-9 images, our validation set will contain 5k samples from Sentinel-2 and 5k samples from Landsat-9).  

We use an equal number of samples per mode when able.  This means that when training with Sentinel-2 and Landsat, each mode has 50K random samples apiece.  When training with Sentinel-1, Sentinel-2, and Landsat, we employ a 20K/40K/40K split.  We take MSELoss per mode, then backpropagate loss through the appropriate alignment head and location encoder.  As such, the overall loss for the location encoder can be thought of as a weighted average of the MSELoss per modality.

We leverage the image augmentations used in SatCLIP's pretraining: random flip, random crop, and Gaussian blur.  For the latitude and longitude, we employ $\sim 1$~km of coordinate jitter.  We pretrain with the RFF \cite{tancik2020fourier} position encoder that was developed for GeoCLIP.  Ablations with the SirenNet  \cite{russwurmgeographic} position encoder backbone can be found in our supplemental materials.  We train on a NVIDIA RTX A6000 GPU using a batch size of 128, a learning rate of 0.001, and an AdamW optimizer.

\begin{table*}[t!]
    \caption{\ours{} results compared to current state-of-the-art models on 4 distinct benchmark suites.  \ours{} uses RFF for position encoding.  Reporting average results $\pm$ standard deviation over 5 randomly seeded runs.  Best results are in \textbf{bold}, second best are \textit{italicized}.  Aggregate results per benchmark suite are at the top of each sub-section}
    \centering
    \begin{tabular}{llllll}
         \toprule Benchmark&SatCLIP&GeoCLIP&\ours&\ours&\ours  \\ 
         \midrule
         \multicolumn{6}{c}{\textbf{Model Details}} \\
         \cmidrule(r){1-6}
         Training Modalities&S2&Flickr&S2&S2+LS&S2+LS+S1\\
         Trainable Parameters&1M&10M&14.1M&14.1M&14.1M\\
         Number of samples&100K&4.7M&100K&100K&100K\\
         Batch size&16K&512&128&128&128\\
         Loss style&InfoNCE&InfoNCE&MSE&MSE&MSE\\
         Hours per epoch &\textit{3.344}&4.912&\textbf{0.105}&\textbf{0.105}&\textbf{0.105}\\
         \midrule
         \multicolumn{6}{c}{\textbf{CHELSA Regression (R$^2$)}} \\
         \cmidrule(r){1-6}
         Aggregate Results&$0.693\pm0.017$&$0.738\pm0.016$&$\textit{0.810}\pm\textit{0.023}$&$\textbf{0.818}\pm\textbf{0.017}$&$0.807\pm0.021$\\
         \midrule
         Frost Change Frequency&$0.651\pm0.008$&$0.658\pm0.023$&$\textit{0.768}\pm\textit{0.017}$&$\textbf{0.780}\pm\textbf{0.020}$&$0.747\pm0.017$\\
         Grow Season Length&$0.770\pm0.021$&$0.779\pm0.010$&$0.863\pm0.018$&$\textit{0.868}\pm\textit{0.003}$&$\textbf{0.879}\pm\textbf{0.011}$\\
         Grow Season First&$0.554\pm0.009$&$0.555\pm0.007$&$\textit{0.680}\pm\textit{0.015}$&$\textbf{0.690}\pm\textbf{0.009}$&$0.671\pm0.020$\\
         Isothermality&$0.757\pm0.013$&$0.847\pm0.006$&$0.941\pm0.004$&$\textbf{0.945}\pm\textbf{0.005}$&$\textit{0.943}\pm\textit{0.005}$\\
         Precipitation Annual&$0.570\pm0.016$&$0.572\pm0.019$&$\textbf{0.694}\pm\textbf{0.027}$&$\textit{0.693}\pm\textit{0.027}$&$0.666\pm0.033$\\
         Precipitation Seasonality&$0.621\pm0.018$&$0.731\pm0.012$&$0.728\pm0.029$&$\textbf{0.754}\pm\textbf{0.014}$&$\textit{0.735}\pm\textit{0.014}$\\
         Snow Cover Days&$0.805\pm0.004$&$0.869\pm0.010$&$\textbf{0.961}\pm\textbf{0.001}$&$\textit{0.959}\pm\textit{0.004}$&$0.958\pm0.003$\\
         Temperature Seasonality&$0.815\pm0.005$&$\textbf{0.867}\pm\textbf{0.002}$&$0.841\pm0.013$&$\textit{0.859}\pm\textit{0.006}$&$0.858\pm0.006$\\
         \midrule
         \multicolumn{6}{c}{\textbf{TorchSpatial Classification (Acc@3)}}\\
         \cmidrule(r){1-6}
         Aggregate Results&$0.888\pm0.001$&$\textbf{0.890}\pm\textbf{0.001}$&$0.888\pm0.001$&$\textbf{0.890}\pm\textbf{0.001}$&$\textbf{0.890}\pm\textbf{0.001}$\\
         \midrule
         iNat2018&$0.904\pm0.001$&$\textit{0.907}\pm\textit{0.000}$&${0.901}\pm0.001$&$\textbf{0.908}\pm\textbf{0.001}$&$0.906\pm0.001$\\
         Flickr&$0.842\pm0.002$&$\textbf{0.844}\pm\textbf{0.002}$&$0.841\pm0.002$&$\textbf{0.844}\pm\textbf{0.001}$&$0.843\pm0.002$\\
         fMoW&$0.917\pm0.001$&$\textbf{0.921}\pm\textbf{0.000}$&$\textbf{0.921}\pm\textbf{0.001}$&$0.918\pm0.001$&$0.920\pm0.001$\\
         \midrule
         \multicolumn{6}{c}{\textbf{TorchSpatial Regression (R$^2$)}} \\
         \cmidrule(r){1-6}
         Aggregate Results&$0.409\pm0.001$&$\textbf{0.503}\pm\textbf{0.001}$&$0.478\pm0.003$&$\textit{0.502}\pm\textit{0.002}$&$0.492\pm0.001$  \\
         \midrule
         Forest Cover&$\textbf{0.675}\pm\textbf{0.000}$&$\textit{0.674}\pm\textit{0.000}$&$0.634\pm0.000$&$0.639\pm0.000$&$0.628\pm0.000$\\
         Elevation&$0.295\pm0.000$&$0.560\pm0.000$&$0.562\pm0.000$&$\textbf{0.641}\pm\textbf{0.000}$&$\textit{0.621}\pm\textit{0.000}$\\
         Nightlights&$0.085\pm0.002$&$\textbf{0.116}\pm\textbf{0.001}$&$0.091\pm0.003$&$\textit{0.100}\pm\textit{0.003}$&$0.092\pm\textit{0.001}$\\
         Population Density&$0.580\pm0.001$&$\textbf{0.661}\pm\textbf{0.001}$&$0.624\pm0.006$&$\textit{0.629}\pm\textit{0.004}$&$0.626\pm0.002$\\
         \midrule
         \multicolumn{6}{c}{\textbf{SustainBench Regression (R$^2$)}}\\
         \cmidrule(r){1-6}
         Aggregate Results&$-0.154\pm0.099$&$\textbf{0.115}\pm\textbf{0.003}$&$0.014\pm0.082$&$0.044\pm0.046$&$\textit{0.077}\pm\textit{0.040}$ \\
         \midrule
         Asset Index&$-0.041\pm0.101$&$\textbf{0.217}\pm\textbf{0.001}$&$0.082\pm0.101$&$0.119\pm0.042$&$\textit{0.161}\pm\textit{0.022}$\\
         Sanitation Index&$-0.193\pm0.100$&$\textbf{0.099}\pm\textbf{0.002}$&$0.015\pm0.073$&$0.059\pm0.032$&$\textit{0.085}\pm\textit{0.054}$\\
         Water Index&$-0.141\pm0.100$&$\textbf{0.036}\pm\textbf{0.007}$&$-0.047\pm0.079$&$0.014\pm0.047$&$\textit{0.029}\pm\textit{0.043}$\\
         Women Education&$-0.241\pm0.135$&$\textbf{0.107}\pm\textbf{0.001}$&$0.005\pm0.108$&$-0.016\pm0.076$&$\textit{0.031}\pm\textit{0.051}$\\
         \bottomrule
         
    \end{tabular}
    
    \label{tab:main_results}
\end{table*}

\subsection{Benchmarking}
Unless explicitly stated, all benchmarks are done on a frozen location encoder (after pretraining as described above) with a trainable linear probe.  We use early stopping while training the linear probe, requiring an improvement of 0.001 in validation loss over 10 epochs in order to continue training.  Training is capped at 100 epochs.  We pretrain 5 randomly seeded runs of each SLED model.  For pre-existing SOTA models, all evaluations are done with 5 randomly seeded runs of the same pretrained model.  All benchmarking is done via simple linear probing in order to explore the latent space with minimal hyperparameter tuning.  

When performing regression, we normalize regression targets to the range of [0,1] using a linear transformation with the lowest and highest available labels.  The only exception is for the TorchSpatial NightLights task, where we use a log transform on the labels instead, as inspired by \cite{klemmerSatCLIPGlobalGeneralPurpose2025}.  These transformations change the scale of the regression targets, but not the distribution.

We evaluate on eight uniformly-at-random (UAR) sampled climate-related variables from the CHELSA-TraCE21k-centennial dataset \cite{karger2023chelsa}, four regression on social indices from SustainBench \cite{yeh2sustainbench},  four regression tasks from TorchSpatial \cite{wu2024torchspatial}, and three classification tasks with fused image embeddings from TorchSpatial \cite{wu2024torchspatial}.  Additional details can be found in the supplemental materials.

We include GeoCLIP \cite{vivancocepedaGeoCLIPClipInspiredAlignment2023} and SatCLIP \cite{klemmerSatCLIPGlobalGeneralPurpose2025} as established benchmarks for location encoders.  While GAIR \cite{liuGAIRImprovingMultimodal2025} and UniGeoCLIP \cite{astrucUNIGEOCLIPUnifiedGeospatial2026} would be intriguing baselines to test against, their model weights have not yet been made publicly available.



\subsection{Downstream Performance}
\label{results:performance}
On CHELSA climate regression tasks, all \ours{} models (regardless of training modality) outperform other location encoders on 7 out of 8 tasks, and ranks second in temperature seasonality (Table \ref{tab:main_results}).  In TorchSpatial classification tasks, \ours{} is on par with other location encoders.  Across TorchSpatial's regression tasks, we see more variance.  \ours{} lags behind existing location encoders in forest cover, but exhibits a sizable boost on elevation mapping, particularly with multimodal versions of SLED.

On the remaining SustainBench regression benchmarks, \ours{} outperforms SatCLIP by a significant margin, but falls short of GeoCLIP.  We believe that this is largely due to these tasks being more observable from ground-level images compared to satellite imagery.  Whereas SatCLIP and \ours{} rely on satellite imagery, GeoCLIP is trained on ground-level images.  However, \ours's performance gain over SatCLIP is noteworthy and further demonstrates the potential of distillation over contrastive learning for location encoding.

We also see the relative training expense in Table \ref{tab:main_results}.  SatCLIP's large batch size and GeoCLIP's large dataset size both act as training bottlenecks in a way that \ours{} is unhindered by, further contextualizing \ours's performance.

\subsection{Modality Ablations}
While pretraining with purely Sentinel-2 imagery results in strong performance for \ours{} as seen in Table \ref{tab:main_results}, pretraining on a mixture of Sentinel-2 and Landsat imagery further improves metrics, particularly on TorchSpatial regression and SustainBench tasks.  

However, in several cases, performance decreases with the inclusion of Sentinel-1 data (when comparing S2+LS vs. S2+LS+S1 variants), perhaps due to fewer channels in Sentinel-1 imagery, which over land has 2 "bands", i.e., polarization VV and VH, compared to 7 spectral bands for Landsat and 12 bands for Sentinel-2.  It is also possible that this decrease is a result of comparatively better multi-spectral (MS) imagery encoders compared to SAR encoders.  Optimal strategies for encoding MS imagery have been explored substantially more compared to SAR data \cite{lane2025genealogy}. Despite this performance dip in some cases, the inclusion of Sentinel-1 provides a boost in model performance on the SustainBench task, while still producing comparable results on CHELSA and TorchSpatial's classification task.


\begin{figure}[t!]
\centering
\begin{minipage}{0.45\textwidth}
  \centering
  \includegraphics[width=\linewidth]{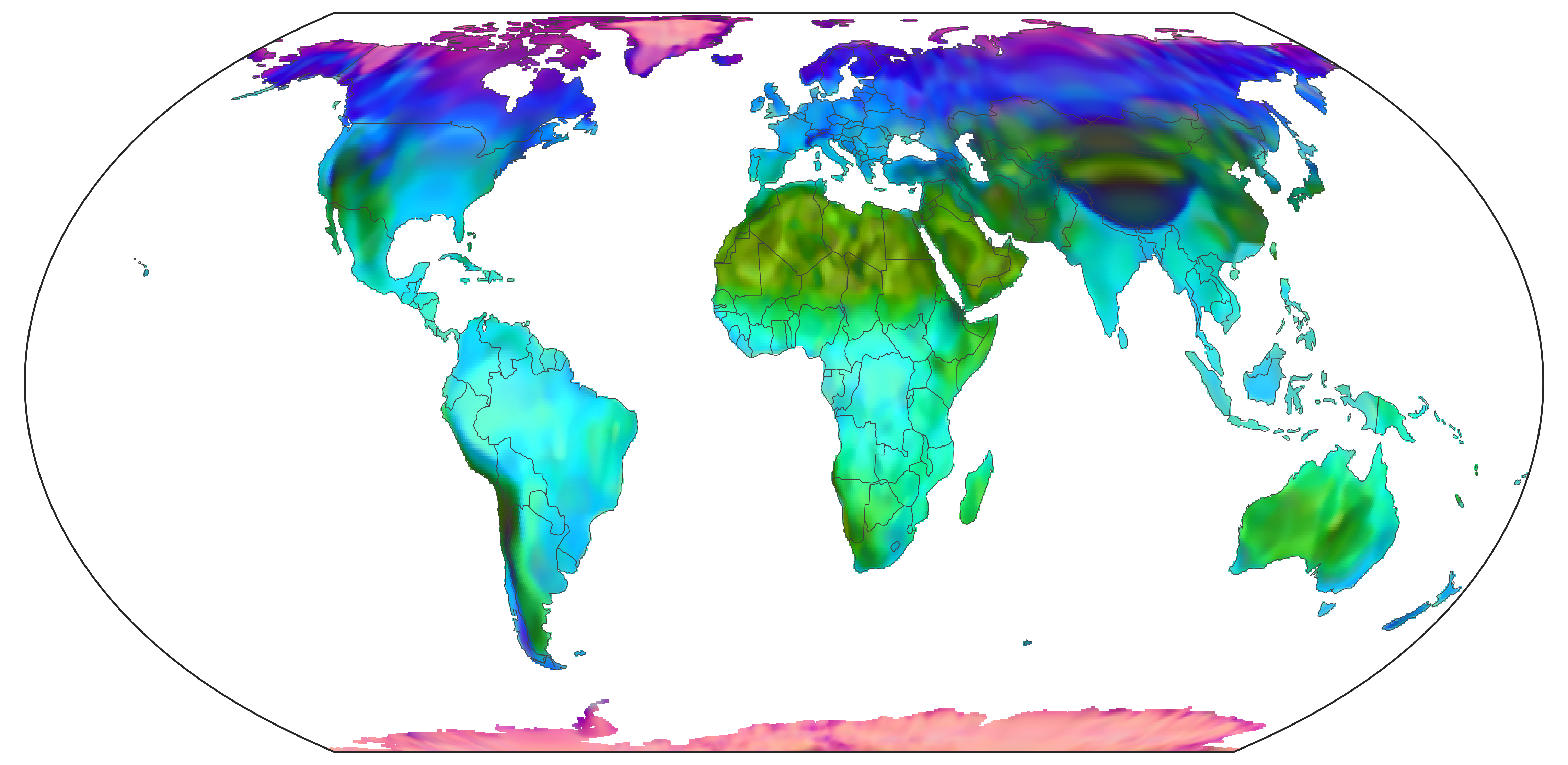}
\end{minipage}%
\hfill
\begin{minipage}{0.45\textwidth}
  \centering
  \includegraphics[width=\linewidth]{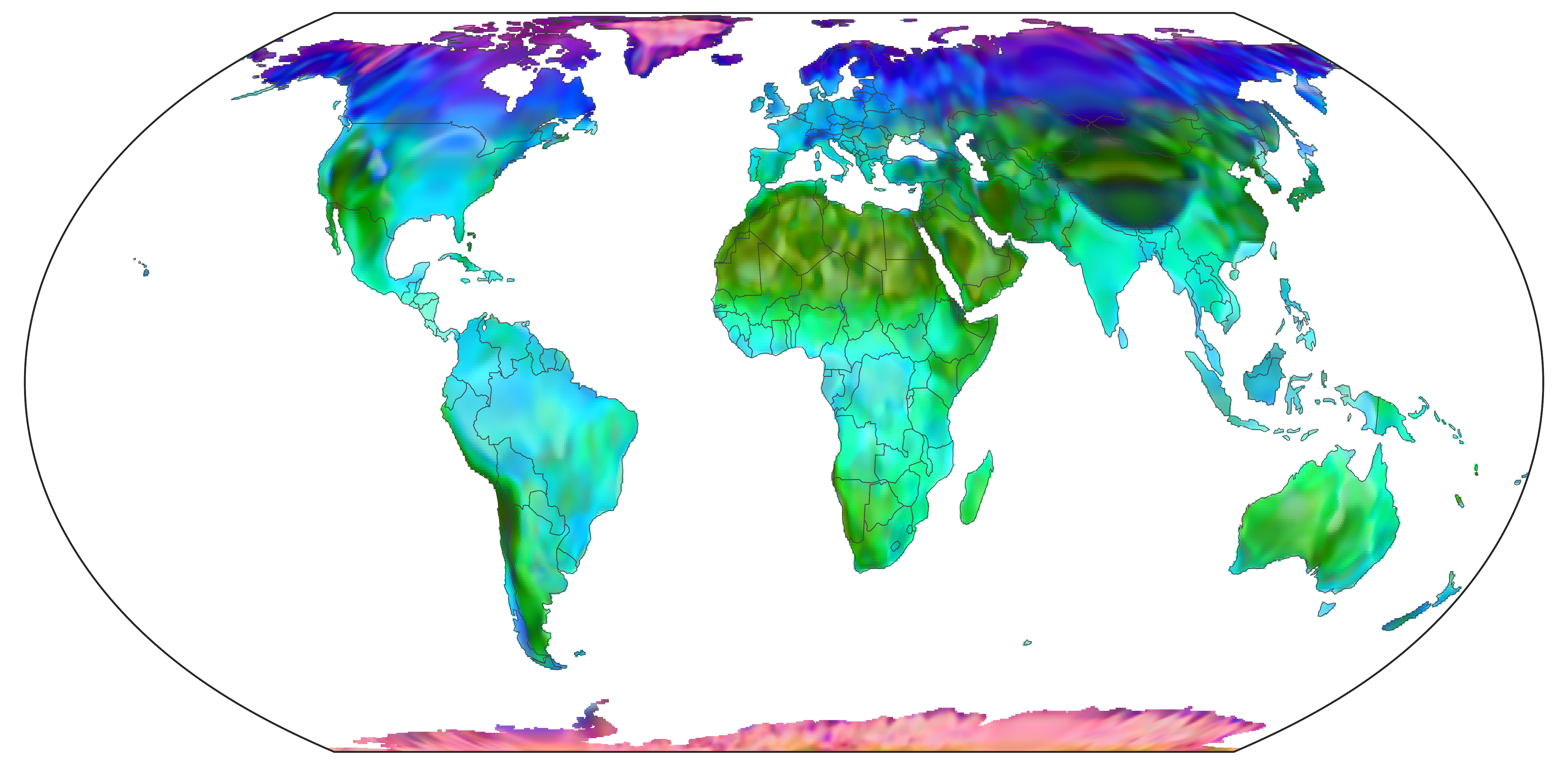}
\end{minipage}%
\hfill
\begin{minipage}{0.45\textwidth}
  \centering
  \includegraphics[width=\linewidth]{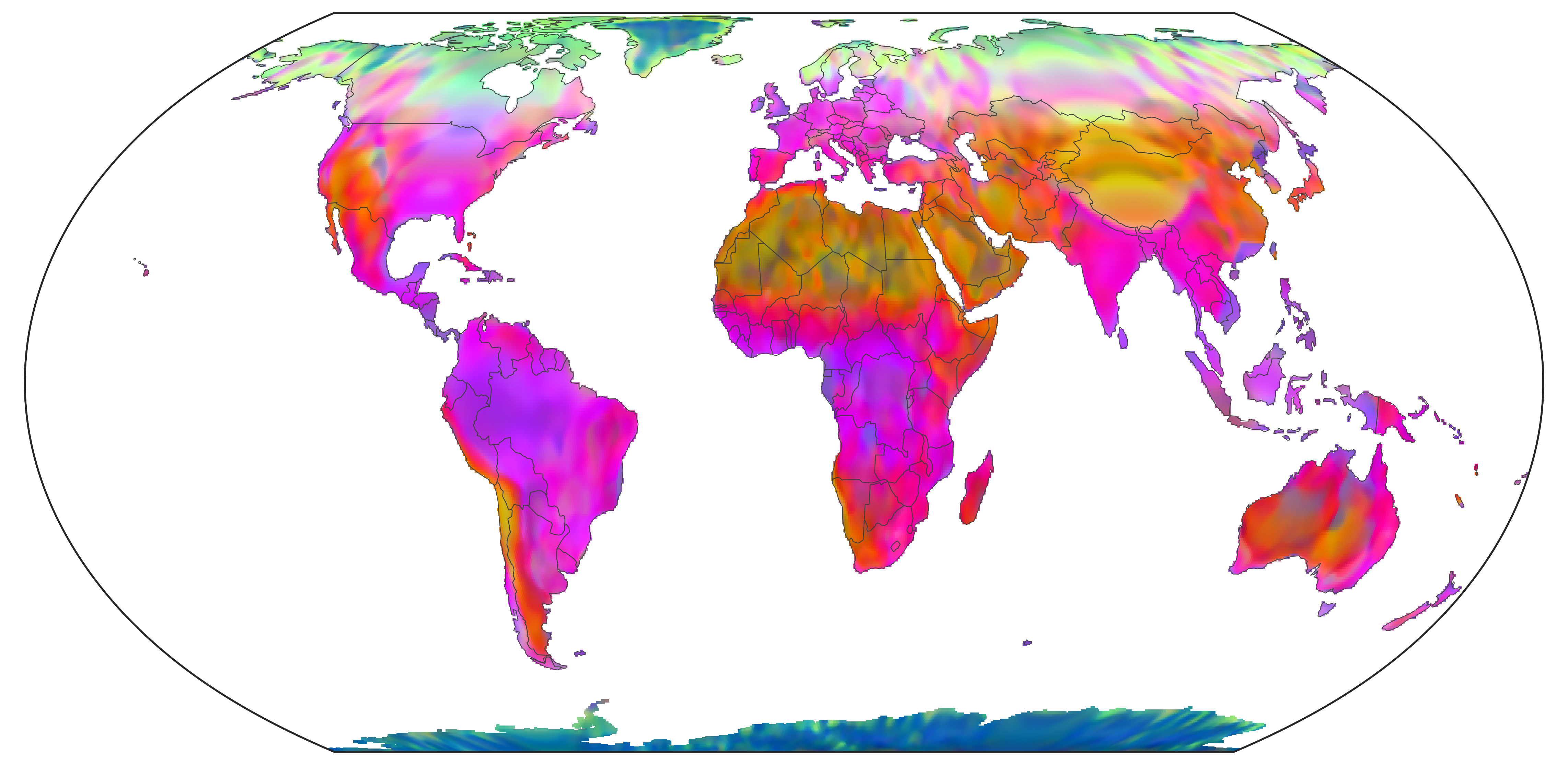}
\end{minipage}
\caption{Equal earth projection of first three principal components of \ours{} embeddings.  From top to bottom, models are \ours{} trained on S2, \ours{} trained on S2+LS and \ours{} trained on S2+LS+S1.  Given the underlying PCA calculations, colors should not be compared from map to map.}
\label{fig:pca_analysis}
\end{figure}


\subsection{Qualitative Embedding Analysis}
For all modality ablations of \ours{},  \ours{} learns meaningful geospatial representations (Figure \ref{fig:pca_analysis}). Embeddings of proximate areas and similar (but distant) biomes are more similar, while the variations within specific regions indicate that \ours{} is also capable of learning region-specific spatial heterogeneity.  The map is also striking in encoding higher elevations, as seen in southwestern China (the Qinghai-Tibet Plateau) and western South America (the Andes).


We also visualize how embeddings change with the addition of new modalities. We use Centered Kernel Alignment (CKA), a latent space similarity metric that is agnostic to model initialization \cite{cortes2012algorithms}. Because CKA needs to be computed over sets of embeddings rather than individual samples, we generate embeddings on a $0.5^\circ$ resolution grid and apply a $3\times3$ sliding window over the resulting embedding maps to define local neighborhoods for CKA analysis. Due to land boundaries and ocean cells, not every grid location contains valid embeddings across all models being compared. To ensure reliable estimates, we only compute CKA scores for center cells with at least three valid spatial locations containing embeddings from both models within the corresponding window.

\begin{figure}[t!]
    \centering
    \includegraphics[width=0.9\linewidth]{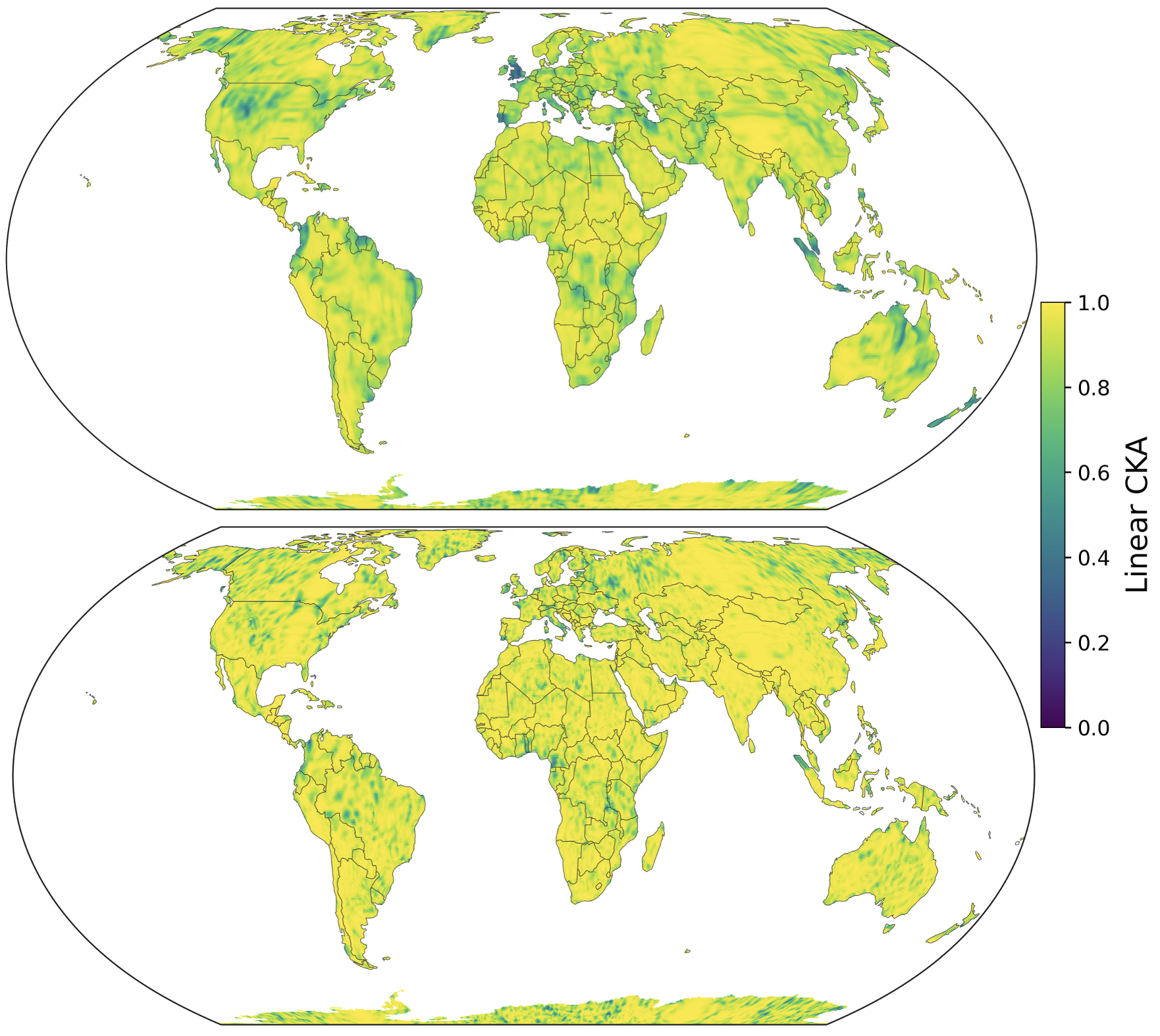}
    \caption{Spatially mapped Centered Kernel Analysis (CKA) of \ours{} embeddings trained on different modalities.  Top:  CKA between \ours{} trained on S2 compared with \ours{} trained on S2/LS.  Bottom: CKA between \ours{} trained on S2+LS compared with \ours{} trained on S2+S1+LS.}
    \label{fig:cka_analysis}
\end{figure}

Within our CKA visualizations, it is notable to see how \ours{} embeddings change from training with Sentinel-2 (which has UAR distribution) to training with Sentinel-2 and Landsat 8/9 (which is from the SSL4EO-L dataset clustered around population centers) in Figure \ref{fig:cka_analysis}.  We see embeddings in more densely populated areas (particularly in the UK, northeast USA, northern South America, southern Africa, and eastern Australia) change the most dramatically.  This demonstrates \ours{}'s ability to leverage modalities with spatially disparate distributions.

Figure \ref{fig:cka_analysis} also demonstrates how adding a notably different training modality (in this case, Sentinel-1 SAR compared with optical imagery from Sentinel-2 and Landsat 8/9) picks up key features that may be missed by other modalities.  We see that embeddings trained with Sentinel-1/Sentinel-2/Landsat differ from optical-only embeddings primarily over water, including the Great Lakes in North America, Lake Victoria in Africa, and bodies of water in eastern Europe.  We include additional CKA and PCA figures in our supplemental materials.

\section{Discussion}
Our results in Table \ref{tab:main_results} demonstrate \ours's ability to capture signals meaningful to a variety of downstream tasks.  We also see in Figure \ref{fig:pca_analysis} that \ours{} is capable of producing geospatially meaningful embeddings, which are the first of its kind to come from a location encoder model trained via distillation. More importantly, we believe that \ours{} will accelerate research in the field, particularly by facilitating the addition of more modes, as each mode only requires a frozen teacher and a simple linear alignment layer.  By leveraging geospatial location as a binding modality, \ours{} only requires the modality-specific teacher and the location in memory at any given time, regardless of the number of modalities used in training.  While the number of frozen parameters will rise with the addition of each mode, the number of trainable parameters needed in memory will remain relatively static.  Mixture of Experts (MoE)-style models serve as a useful precedent here, as they separate performance from computation requirements by limiting the number of trainable parameters necessary to have in memory at any given time \cite{gamalMoEScaleModular2025}.  As such, we think \ours{} could further facilitate multimodal training for location encoders.

The teacher-based formulation of \ours{} enables flexible choices in how modalities are incorporated.  We currently train with a teacher per modality, but the rise of multimodal remote sensing encoders means that a single multimodal teacher, such as TerraFM \cite{danish2025terrafm}, DOFA \cite{xiong2024neural}, OmniSat \cite{astrucOmniSatSelfsupervisedModality2025}, could serve as the teacher for a number of remotely sensed modes, cutting down the number of frozen parameters required in the model.  Conversely, one could also leverage multiple different teachers for a single mode.  Our modality ablations investigate the common case where each additional mode is introduced through a dedicated teacher, but the framework naturally supports decoupling teacher and modality scaling.  Recent work shows that adding an additional teacher for distillation to existing pretraining frameworks for satellite \textit{image} encoders improves performance \cite{mendieta2023towards}, suggesting that additional teachers serve as regularizers in pretraining. Therefore, scaling teacher diversity alongside modality diversity is a promising direction for multimodal location encoding.


Our results suggest that the addition of modalities generally improves model performance, with the largest gains from information-rich multi-spectral imagery. Across several tasks, multimodal variants of \ours{} outperform single-mode models. The modality ablations in Table~\ref{tab:main_results} and latent space analysis in Figure~\ref{fig:cka_analysis} further demonstrate that \ours{} can incorporate additional modalities while maintaining the expressive structure learned from the original Sentinel-2 modality. Although the benefits of adding Sentinel-1  SAR are less consistent than we expected, this may be caused by fewer channels in Sentinel-1 SAR imagery when compared to Sentinel-2 and Landsat MS imagery. The value of incorporating SAR may also become more apparent as SAR-specific benchmark tasks become available.

It is also worth addressing the metrics observed on the SustainBench suite. GeoCLIP performs best on these tasks, which we hypothesize is due to the benchmark containing tasks with stronger relationships to observable landscape characteristics. However, all models perform relatively poorly on this benchmark, where targets are primarily derived from survey-based measurements. 



\section{Conclusion}

In this paper, we presented \ours{}, a distillation-based framework for location encoding. Our unimodal experiments demonstrate that distillation is a viable strategy for training location encoders more quickly with smaller batch sizes. We further showed how this framework extends to multimodal learning by using geospatial location as a binding modality, eliminating the need for costly spatiotemporal alignment (i.e., coregistration). Both unimodal and multimodal \ours{} achieve competitive performance with SOTA location encoders while providing a scalable, efficient approach for multimodal geospatial representation learning.

\section*{Acknowledgments}

This research was supported by the CU Boulder Research \& Innovation Office.

Many thanks to those in the GeoHAI Lab and Rolf Lab for their kindness and support.

\bibliography{references}


\section{Supplemental Materials}
\subsection{Technical Details}
\subsubsection{Pretraining}
\label{supplement:unimodal training}
When training \ours{} on Sentinel-2 imagery, we use all 100K samples from the S2-100K dataset and use a frozen TerraFM \cite{danish2025terrafm} encoder to encode the images.  We feed the latitude and longitude through a trainable position encoder, then compute a simple MSELoss between the position embeddings and the image embeddings.  The resulting loss is backpropagated through the position encoder.  We can think of the frozen TerraFM encoder as the teacher and the position encoder as the student in this distillation setup.  Since TerraFM embeds 768 dimensions, our position encoder also encodes in 768 dimensions.  We utilize TerraFM since it trains on Sentinel-2's ground-level L2A imagery \cite{spotoOverviewSentinel22012}, which is the data product available in the S2-100K dataset \cite{klemmerSatCLIPGlobalGeneralPurpose2025}.  Most other Sentinel-2 encoders train on the raw Sentinel-2 L1C data \cite{lane2025genealogy}.

When training on Landsat imagery, we use a frozen ViT-Base MoCo \cite{he2020momentum} encoder, made available through TorchGeo \cite{stewart2024torchgeo}, trained on Landsat OLI SR imagery as the teacher model.  We randomly sample the SSL4EO-L dataset to obtain our 100K samples.  Relatively few GeoFMs are trained only on Landsat data \cite{lane2025genealogy}.  We leverage this MoCo model as a way to have a unimodal teacher for Landsat, as opposed to a teacher that will generate multimodal embeddings, such as TerraMind \cite{jakubik2025terramind} or OmniSat \cite{astrucOmniSatSelfsupervisedModality2025}.

When training on Sentinel-1 imagery, we use the 20k samples from our custom Sentinel-1 dataset, which is partially available in our code supplement.  We use the FG-MAE \cite{wang2024feature} for encoding SAR imagery, as we found the feature-guided reconstruction targets a particularly compelling SSL task for generating robust representations of Sentinel-1 imagery.

When training multimodal \ours{}, we use PyTorch Lightning's CombinedDataLoader class to load in several modality-specific dataloaders and employ the max size combination strategy.  This means that every batch will contain K modes until we have exhausted all the training samples for one mode.  Then every batch will contain K-1 modes, and so on.  Our training code can be found in the code supplement; the ReadMe.md file should contain appropriate details regarding the repository.

\subsubsection{Benchmark Suites}
\label{supplement:benchmarks}
We evaluate on 8 distinct climate-related variables as regression targets.  This benchmark consists of 10K locations, sampled in UAR across all landmasses on Earth.  The regression targets are annual temperature, annual precipitation, temperature seasonality, precipitation seasonality, isothermality, snow cover days, frost change frequency, first day of growing season, and growing season length.  All variables in question are taken from the CHELSA-TraCE21k-centennial dataset \cite{karger2023chelsa}, which contains 1km-resolution maps of all variables.  All tasks use a random train/val/test split of 60\%/20\%/20\%.  This dataset can be found in our code supplement.

We also benchmark against 4 regression tasks (forest cover, elevation, nightlights, and population density) from TorchSpatial \cite{wu2024torchspatial}.  Lastly, we include an additional 3 classification tasks, also with fused image embeddings, from TorchSpatial \cite{wu2024torchspatial}: iNat2018 \cite{van2018inaturalist}, YFCC \cite{tang2015improving}, and Functional Map of the World \cite{christie2018functional}. Specifically, we concatenate the pretrained Inception-V3 image features from the benchmark with location embeddings, then pass the combined vector through a final classification layer. Lastly, we include 4 regression tasks from SustainBench \cite{yeh2sustainbench}: asset index, water index, sanitation index, and women education index.  While SustainBench makes ground-level Mapillary \cite{neuhold2017mapillary} imagery, composite Landsat imagery, and nightlights imagery available for each sample, we simply regress against the latitude and longitude to benchmark models.  This accounts for the difference seen between the results reported in \cite{yeh2sustainbench} and our paper.

For all benchmarks, we use a simple linear layer as our classification head.  This differs from other works \cite{klemmerSatCLIPGlobalGeneralPurpose2025, wu2024torchspatial}, which have used a trainable MLP as their projection head, and accounts for the discrepancy between our results and theirs.  We use a linear layer as a means to avoid costly hyperparameter tuning and ensure that benchmarking is a function of the latent space rather than a specific classification head architecture.

\subsection{Ablations}
\begin{table}[]
    \centering
    \caption{Ablations between \ours{} location encoder using Spherical Harmonics, Random Fourier Features, SirenNet, and UniGeoCLIP's multi-scale RFF encoder.  Examining models trained with Sentinel-1 (S1), Sentinel-2 (S2), and Landsat 8/9 (LS).  Results contain mean plus/minus standard deviation across 5 randomly seeded runs.  Best results are in \textbf{bold} and second-best are in \textit{italics}.}
    \label{tab:siren_net}
    \begin{tabular}{llll}
    \toprule
     Modalities&\shortstack{CHELSA \\(R$^2$)}&\shortstack{SustainBench \\(R$^2$)}&\shortstack{eRank\\(dim)}\\
     \midrule
     \multicolumn{4}{c}{Random Fourier Features} \\
     \midrule
     S2&$\textit{0.81}\pm\textit{0.02}$&$0.01\pm0.08$&$35.4$ \\
     S2+LS &$\textbf{0.82}\pm\textbf{0.02}$&$\textit{0.04}\pm\textit{0.05}$&$48.9$ \\
     S2+LS+S1&$0.81\pm0.02$&$\textbf{0.08}\pm\textbf{0.04}$&$45.7$ \\
     \midrule
     \multicolumn{4}{c}{Spherical Harmonics} \\
     \midrule
     S2&$-0.53\pm0.42$&$-2.03\pm0.96$&$22.4$\\
     S2+LS&$-0.59\pm0.28$&$-1.66\pm0.52$&$42.1$\\
     S2+LS+S1&$-0.75\pm0.37$&$-1.94\pm0.79$&$40.4$\\
     \midrule
     \multicolumn{4}{c}{SirenNet} \\
     \midrule
     S2+LS+S1&$-0.15\pm0.23$&$-1.79\pm0.48$&$\textit{48.9}$\\
     \midrule
     \multicolumn{4}{c}{UniGeoCLIP} \\
     \midrule
     S2+LS+S1&$0.73\pm0.02$&$-1.07\pm0.55$&$\textbf{57.8}$\\
     \bottomrule

    \end{tabular}
    
\end{table}
\subsubsection{Position Encoders}
\label{supplement:siren_net}
We trained additional models with SirenNet \cite{sitzmann2020implicit}, Spherical Harmonics \cite{russwurmgeographic} and UniGeoCLIP's multi-scale RFF encoder \cite{astrucUNIGEOCLIPUnifiedGeospatial2026} as a position encoder, but the results lagged behind the RFF encoder used in GeoCLIP \cite{vivancocepedaGeoCLIPClipInspiredAlignment2023}, often dramatically. Given the lack of performance, we ceased using compute resources to evaluate non-RFF models.  However, we do have results available for 5 randomly seeded runs across all CHELSA and SustainBench tasks.  When training \ours{} with Spherical Harmonics, we use 10 Legendre Polynomials.  All variations of \ours{} encode in 768 dimensions.  For RFF, SirenNet, and Spherical Harmonics, we use the default constructors available through the RSHF python package \cite{RSHF2024}.  The details for our encoder variations for \ours{} are available here in Supplement table \ref{tab:siren_net}.  All models training on more than one mode use the alignment head method detailed in the main paper.  All benchmarking is done via linear probing.

We find that the RFF encoder used in GeoCLIP \cite{vivancocepedaGeoCLIPClipInspiredAlignment2023} is by far the best performing position encoder option.  The UniGeoCLIP multi-scale RFF encoder comes reasonably close on the CHELSA suite, but falls behind in the SustainBench suite.  We also include the effective rank (eRank) \cite{roy2007erank} of each model as a way to quickly capture how many dimensions in our latent space have meaningful variation.  In other words, eRank provides a useful tool for demonstrating how much of our latent space is effectively being used with each approach.  We see how the addition of Landsat as a modality results in a meaningful improvement in eRank for both RFF and Spherical Harmonics versions of \ours.  We also see eRank dip slightly with the inclusion of Sentinel-1, which we theorize is due to the relatively fewer number of bands in SAR imagery when compared to multispectral imagery from Sentinel-2 or Landsat.

Interestingly, we see eRanks that are comparable to (or better than) RFF versions of \ours{} for \ours{} using SphericalHarmonics, SirenNet, and UniGeoCLIP.  This suggests that, despite their currently poor downstream performance, there is untapped potential in leveraging additional position encoding strategies for distillation-style location encoders.


    

\begin{table}[]
    \centering
    \caption{Examining cross-modal alignment options performance.  All models train with Sentinel-1, Sentinel-2, and Landsat 8/9, with a 20/40/40 training dataset split.  Options are Projection Head (Project.), Matryoshka embedding (Matryo.), or Alignment Heads (Align.).  Best results are in \textbf{bold}, second best are \textit{italicized}.}
    \label{tab:alignment}
    \begin{tabular}{llll}
    \toprule
    Method&\shortstack{Biome \\(acc)}& \shortstack{EcoRegion \\(acc)}&\shortstack{Temp \\($R^2$)}\\
    \midrule
    \multicolumn{4}{c}{Random Fourier Features} \\
    \midrule
     Project.&$0.72\pm0.00$&$0.74\pm0.00$&$-0.01\pm0.00$  \\
     Matryo.&$\textit{0.76}\pm\textit{0.01}$&$\textit{0.74}\pm\textit{0.01}$&$\textit{0.79}\pm\textit{0.78}$\\
     Align.&$\textbf{0.77}\pm\textbf{0.00}$&$\textbf{0.76}\pm\textbf{0.00}$&$\textbf{0.88}\pm\textbf{0.01}$\\
     \midrule
     \multicolumn{4}{c}{Spherical Harmonics} \\
     \midrule
     Project.&$0.71\pm0.01$&$0.72\pm0.01$&$-0.02\pm0.01$\\
     Matryo.&$0.75\pm0.02$&$0.63\pm0.04$&$-0.07\pm0.22$\\
     Align.&$0.74\pm0.02$&$0.56\pm0.06$&$-0.46\pm0.32$\\
     \bottomrule
    \end{tabular}
    
\end{table}

\subsubsection{Cross-Modal Alignment}
\label{supplement:framework}
We explored a variety of options for how to calculate loss across multimodal teachers with different latent space sizes.  Given that GeoFMs have any number of different embedding dimensions, ensuring that \ours{} could easily incorporate new, potentially disparate GeoFMs, was important.

We examined 3 options for aligning latent spaces before calculating loss:
\begin{enumerate}
    \item Projection head: adding a trainable projection head atop each modality specific teacher.  This projection head would be a linear layer that projected the teacher to the same number of dimensiosn as \ours's latent space.
    \item Matryoshka: a form of Matryoshka embeddings \cite{kusupatiMatryoshkaRepresentationLearning2024}, where we perform MSE loss on the first N available dimensions, where N is the minimum of the number of \ours's dimensions in the latent space and number of the teacher's  dimensions in the latent space.
    \item Alignment heads: the alignment head method described in the main paper.  This differs from projection heads in that the alignment heads sit atop the location encoder, \textit{not} the modality specific teachers.  This means that the loss for our alignment heads is propagated within the context of the rest of our trainable parameters in the location encoder.
\end{enumerate}

We examine these methods on three benchmarks: Koppen biome classification \cite{kottek2006world}, EcoRegion classification \cite{dinerstein2017ecoregion}, and temperature regression from CHELSA.  Results are averaged over 3 runs and can be found in Supplement Table \ref{tab:alignment}.  We see that projection heads and Matryoshka loss both lag behind our preferred approach of using alignment heads.

\section{Additional Figures}
\begin{figure*}[t!]
    \centering

    \begin{minipage}{\linewidth}
        \centering
        \includegraphics[width=\linewidth]{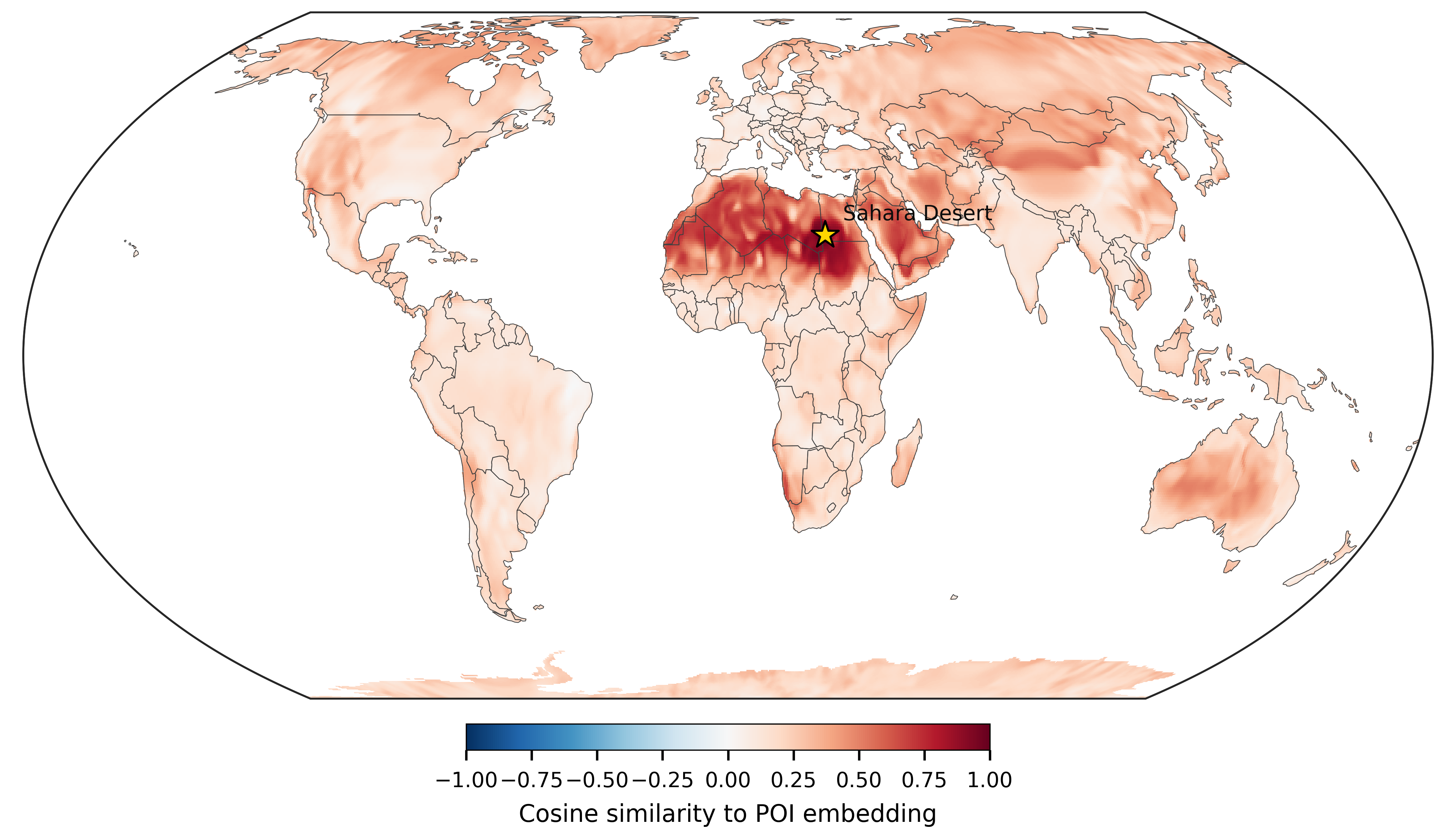}
    \end{minipage}
    \begin{minipage}{\linewidth}
        \centering
        \includegraphics[width=\linewidth]{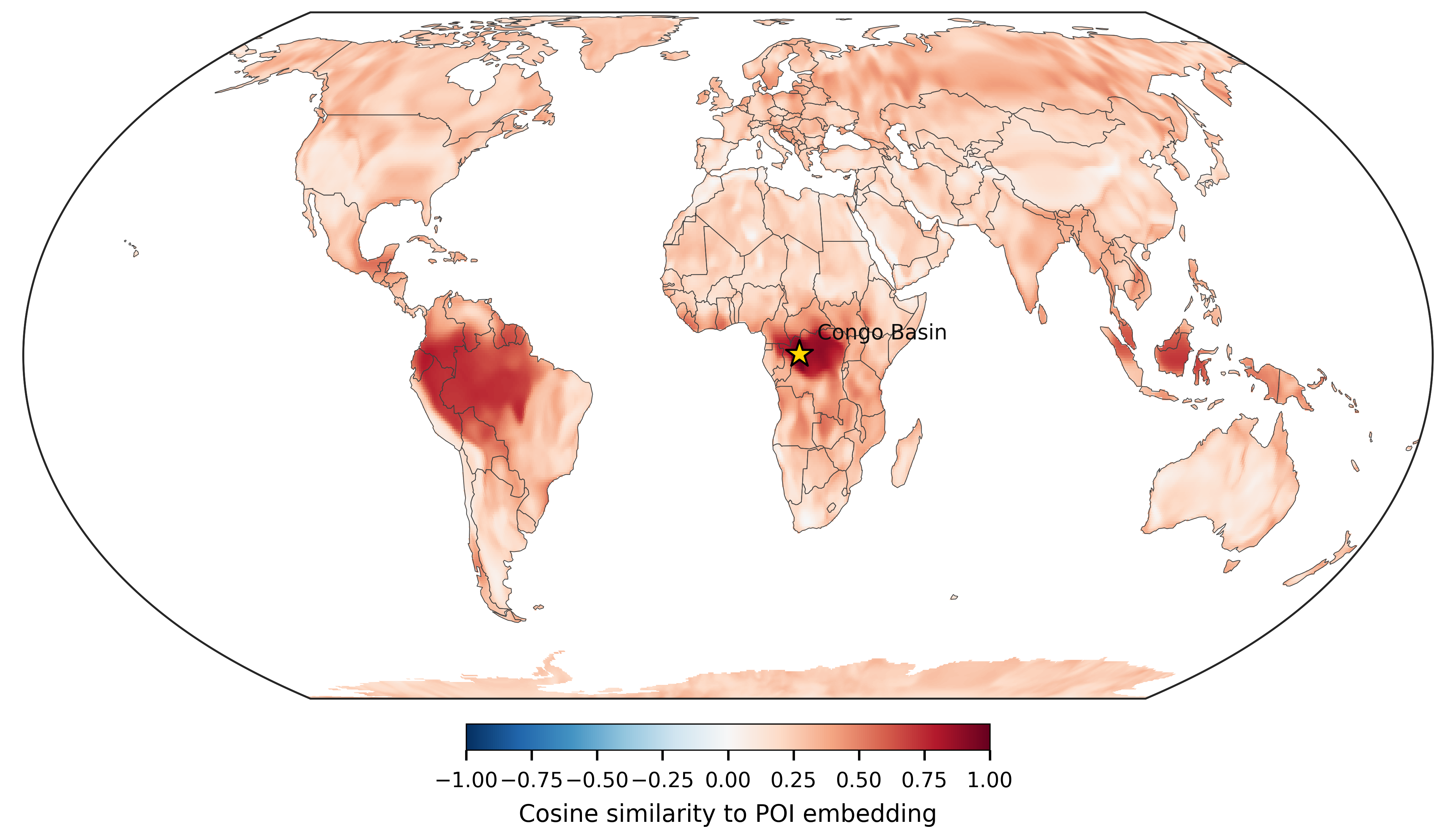}
        
    \end{minipage}

    \caption{Spatial patterns of cosine similarity between global location embeddings and selected point-of-interest (POI) embeddings. The top figure shows a POI located in the Sahara Desert, while the bottom shows a POI located in the Congo Basin.  The resulting similarity maps show that \ours{} captures geographically and environmentally meaningful relationships, with high similarity concentrated around regions sharing similar spatial or environmental characteristics with each reference location.}
    \label{fig:cos_sim}
\end{figure*}

We provide additional map visualizations in the hopes that these can provide qualitative understanding of \ours.  In Figure \ref{fig:cos_sim}, we see embedding cosine similarity compared to a single point of interest.  In both instances, we see the surrounding region and other similar biomes around the world highlighted as the most similar.

The following pages include CKA and PCA visualizations already seen in the main paper, but blown up in size for convenient reference.

\begin{figure*}[]
    \centering
    \includegraphics[width=1.0\linewidth]{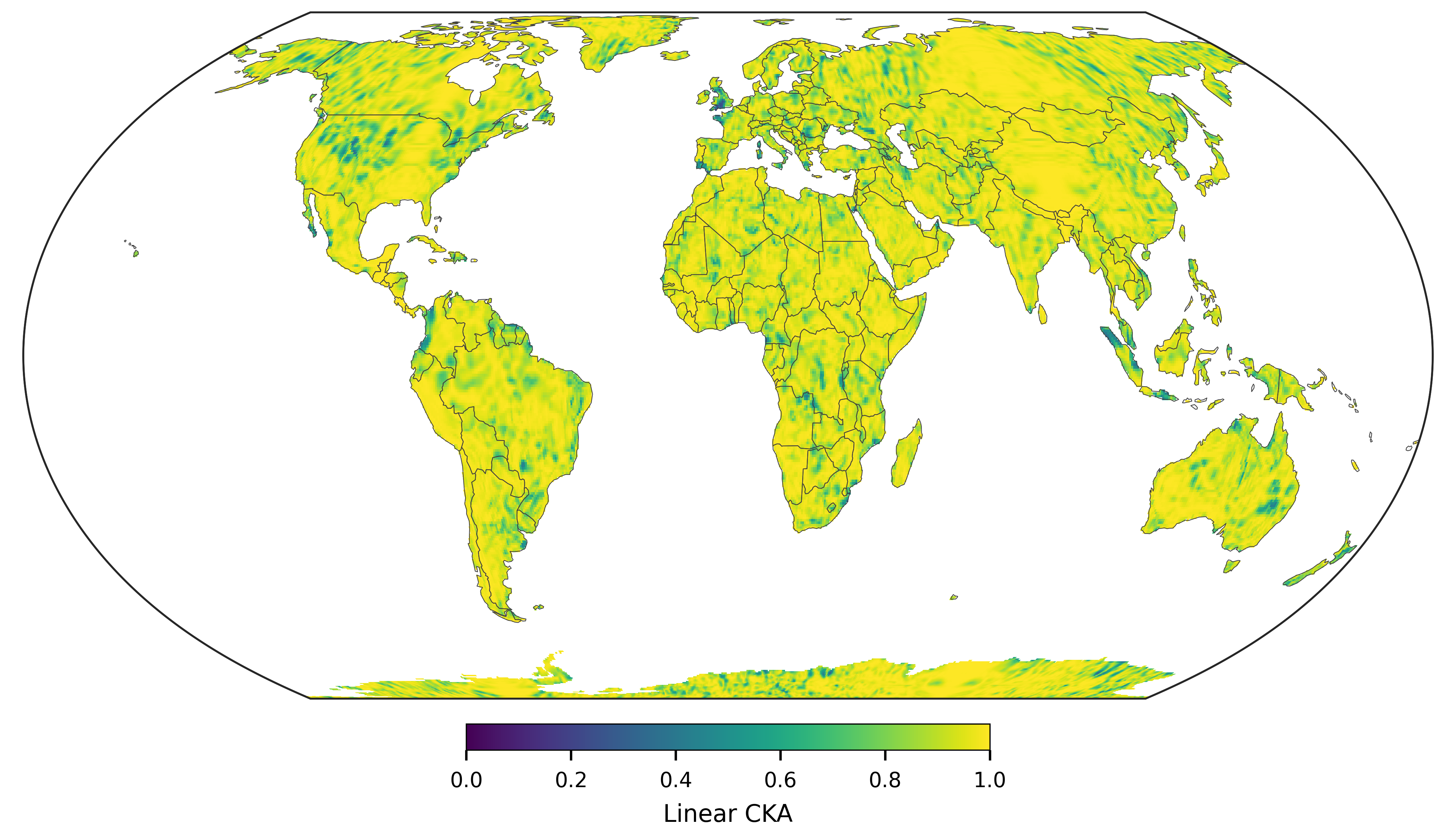}
    \caption{Blown-up version of CKA analysis of Sentinel-2 \ours{} compared with Sentinel-2/Landsat \ours.}
    \label{fig:cka_landsat_large}
\end{figure*}

\begin{figure*}[]
    \centering
    \includegraphics[width=1.0\linewidth]{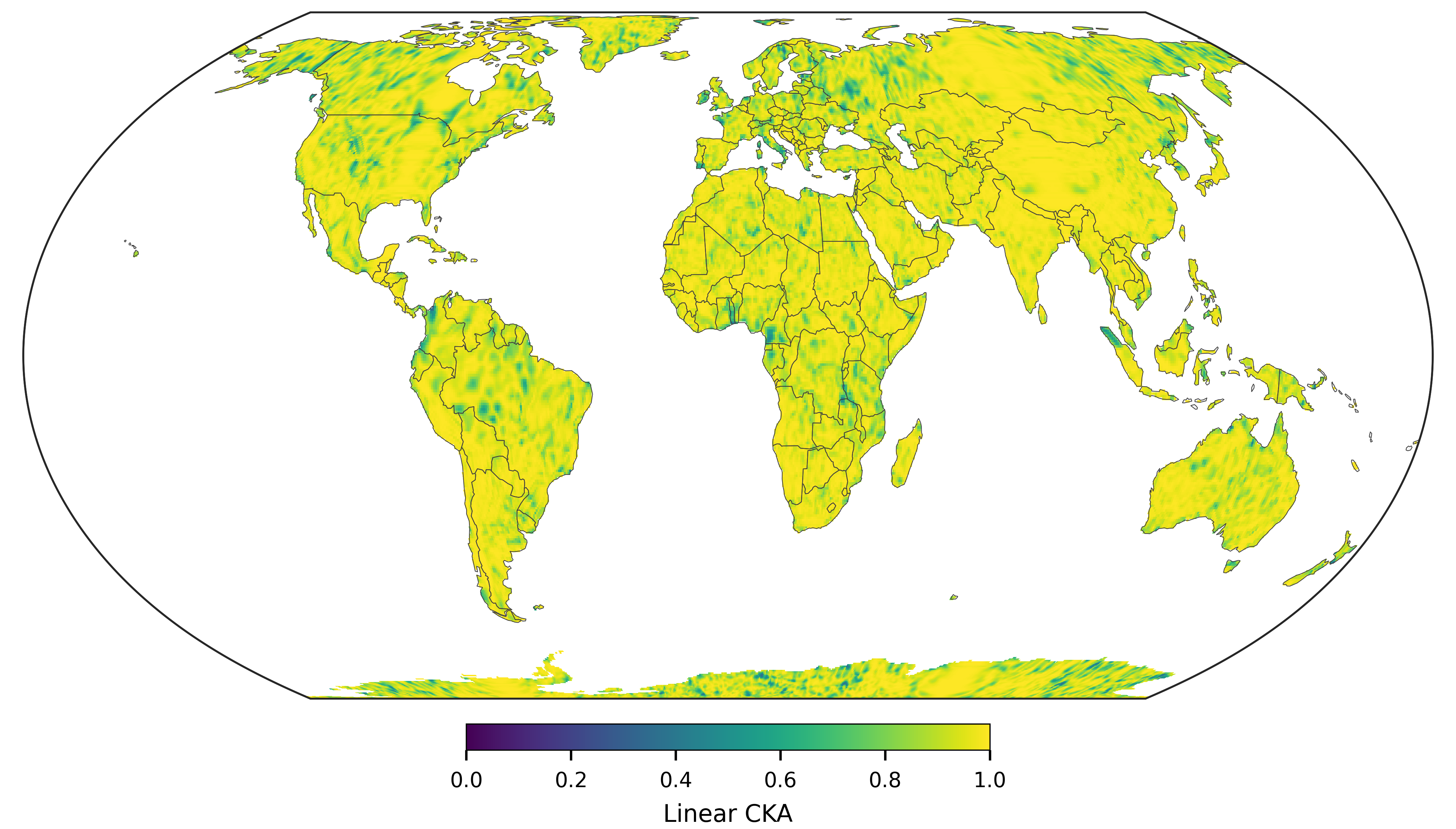}
    \caption{Blown up CKA analysis of Sentinel-2/Landsat \ours{} compared with Sentinel-1/Sentinel-2/Landsat \ours.}
    \label{fig:cka_sentinel_1}
\end{figure*}

\begin{figure*}[]
    \centering
    \includegraphics[width=1.0\linewidth]{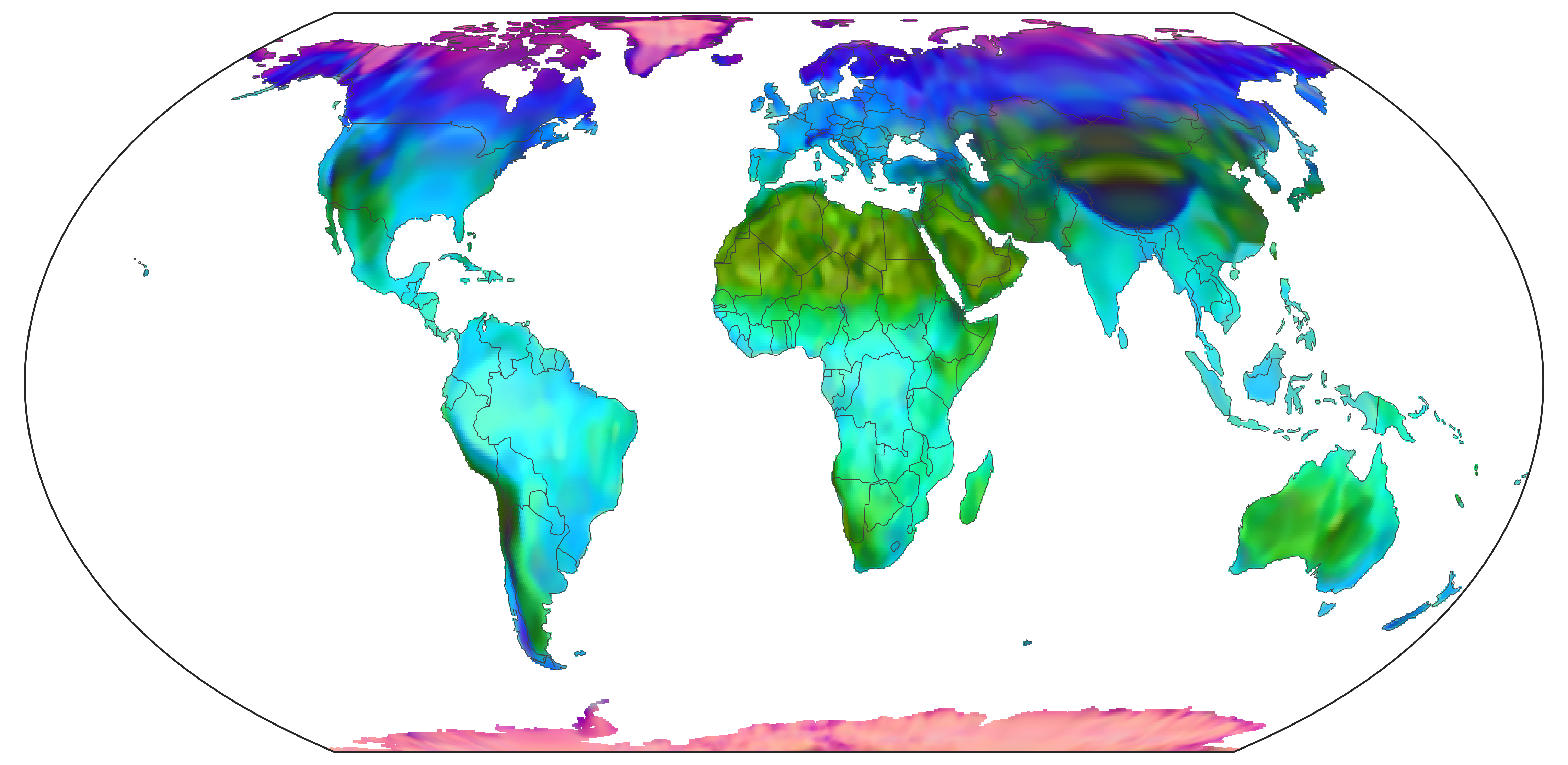}
    \caption{Blown-up version of first three principal components visualization of \ours{} embeddings trained on Sentinel-2.}
    \label{fig:pca_sled_s2large}
\end{figure*}

\begin{figure*}[]
    \centering
    \includegraphics[width=1.0\linewidth]{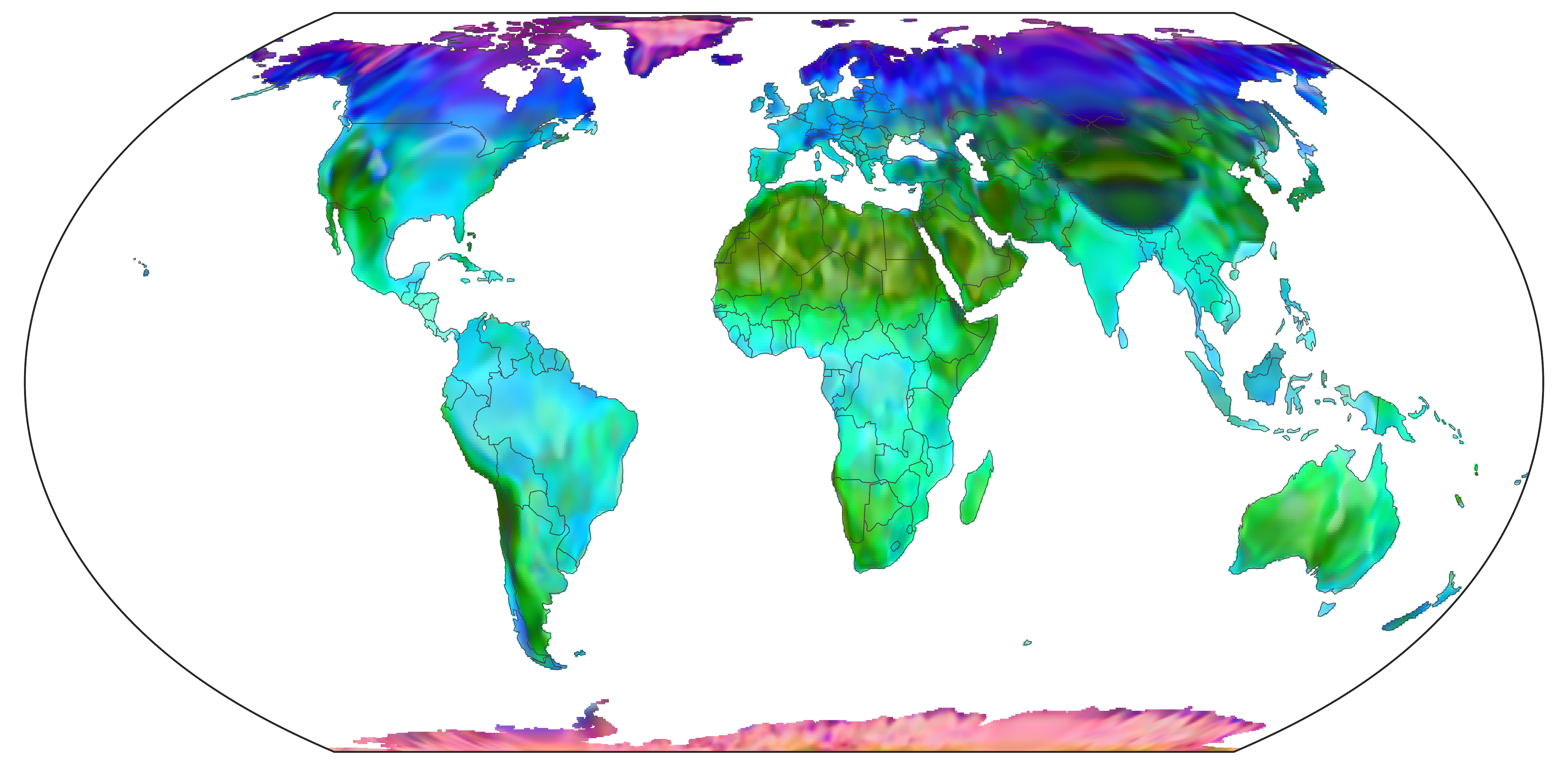}
    \caption{Blown-up version of first three principal components visualization of \ours{} embeddings trained on Sentinel-2 and Landsat 8/9.}
    \label{fig:pca_sled_s2_ls_large}
\end{figure*}

\begin{figure*}[]
    \centering
    \includegraphics[width=1.0\linewidth]{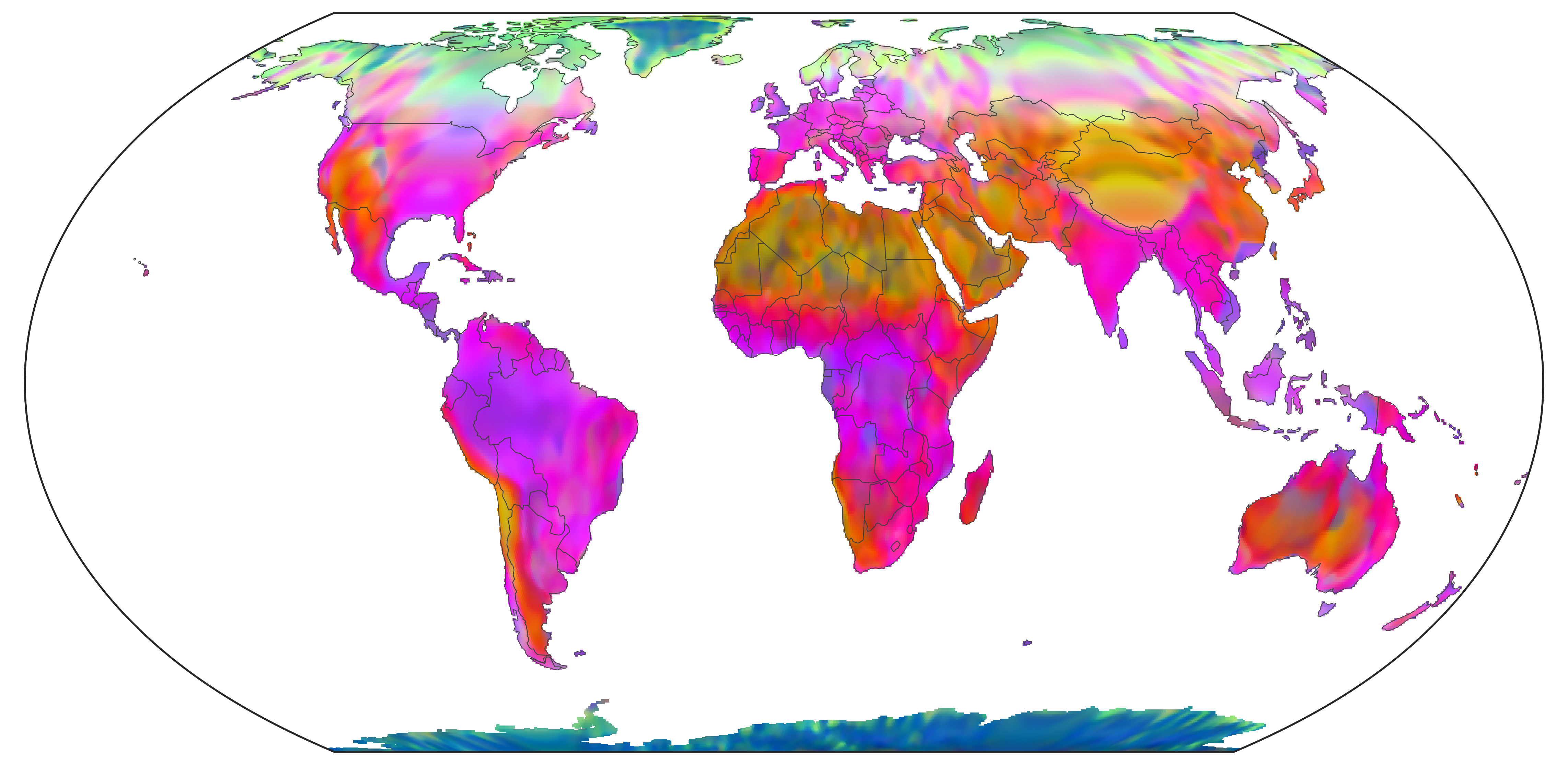}
    \caption{Blown-up version of first three principal components visualization of \ours{} embeddings trained on Sentinel-2, Landsat 8/9, and Sentinel-1.  To be clear, the colors of PCA are individualized per map.  Colors in this figure should not be compared with \ref{fig:pca_sled_s2_ls_large} or \ref{fig:pca_sled_s2large}}
    \label{fig:pca_sled_s12_ls_large}
\end{figure*}

\end{document}